\documentclass[lettersize,journal]{IEEEtran}
\usepackage{amsmath,amsfonts}
\usepackage{algorithmic}
\usepackage{algorithm}
\usepackage{array}
\usepackage{textcomp}
\usepackage{stfloats}
\usepackage{url}
\usepackage{verbatim}
\usepackage{graphicx}
\usepackage{cite}
\usepackage{siunitx}
\usepackage[T1]{fontenc} 

\usepackage{etoolbox}
\makeatletter
\patchcmd{\@makecaption} 
  {\scshape}
  {}
  {}
  {}
\patchcmd{\@makecaption} 
  {\\}
  {.\ }
  {}
  {}
\patchcmd{\@makecaption} 
  {\centering}
  {}
  {}
  {}
\makeatother

\begin{document}

\title{Metrically Scaled Monocular Depth Estimation through Sparse Priors for Underwater Robots}

\author{
    Luca Ebner$^1$,
    Gideon Billings$^2$,
    Stefan Williams$^2$
    \thanks{$^1$ Luca Ebner is with the Robotic Systems Lab, ETH Zurich and with the Australian Centre for Field Robotics, University of Sydney {\tt ebnerl@ethz.ch}}
    \thanks{$^2$ Gideon Billings and Stefan Williams are with the Australian Centre for Field Robotics, University of Sydney}
} 



\maketitle

\begin{abstract}
In this work, we address the problem of real-time dense depth estimation from monocular images for mobile underwater vehicles. We formulate a deep learning model that fuses sparse depth measurements from triangulated features to improve the depth predictions and solve the problem of scale ambiguity. 
To allow prior inputs of arbitrary sparsity, we apply a dense parameterization method.
Our model extends recent state-of-the-art approaches to monocular image based depth estimation, using an efficient encoder-decoder backbone and modern lightweight transformer optimization stage to encode global context. The network is trained in a supervised fashion on the forward-looking underwater dataset, FLSea. Evaluation results on this dataset demonstrate significant improvement in depth prediction accuracy by the fusion of the sparse feature priors. In addition, without any retraining, our method achieves similar depth prediction accuracy on a downward looking dataset we collected with a diver operated camera rig, conducting a survey of a coral reef. The method achieves real-time performance, running at 160 FPS on a laptop GPU and 7 FPS on a single CPU core and is suitable for direct deployment on embedded systems. The implementation of this work is made publicly available at \url{https://github.com/ebnerluca/uw_depth}.
\end{abstract}

\begin{IEEEkeywords}
Marine Robotics, Computer Vision for Automation, Deep Learning for Visual Perception
\end{IEEEkeywords}

\section{Introduction}
\label{sec:introduction}

\begin{figure}[h]
    \centering
    \begin{minipage}{0.32\linewidth}
        \centering
        \includegraphics[width=\linewidth]{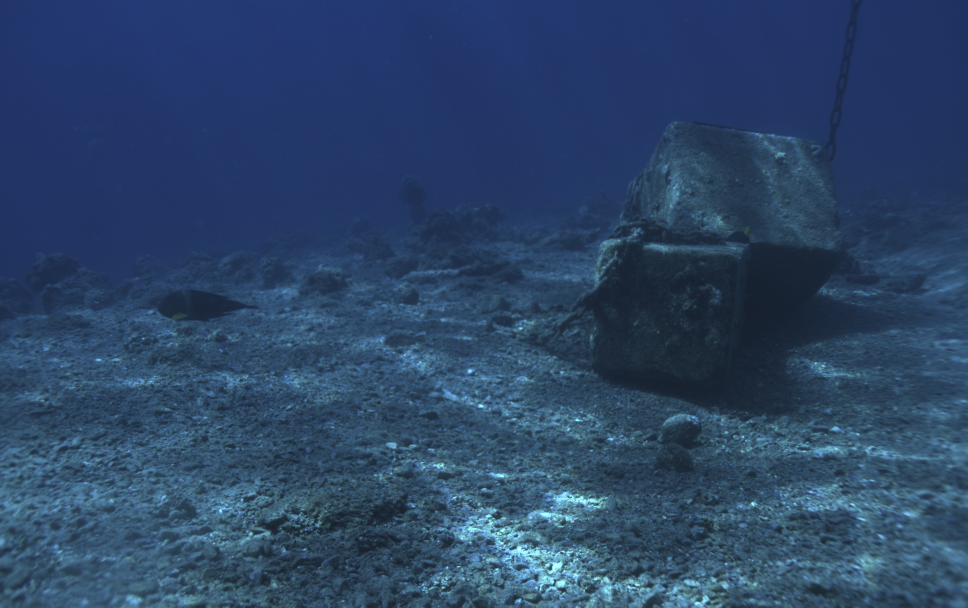}%
    \end{minipage}
    \begin{minipage}{0.32\linewidth}
        \centering
        \includegraphics[width=\linewidth]{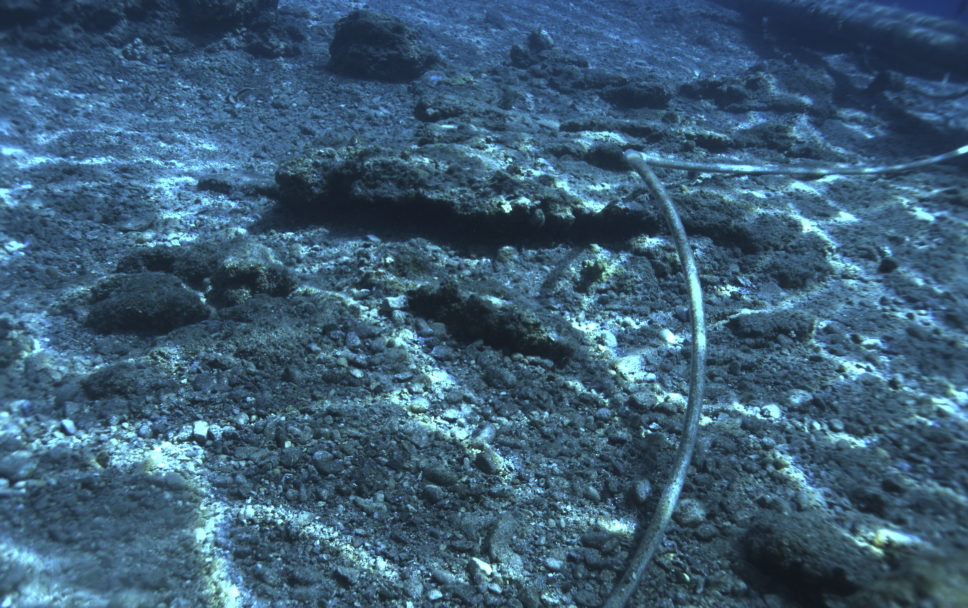}%
    \end{minipage}
    \begin{minipage}{0.32\linewidth}
        \centering
        \includegraphics[width=\linewidth]{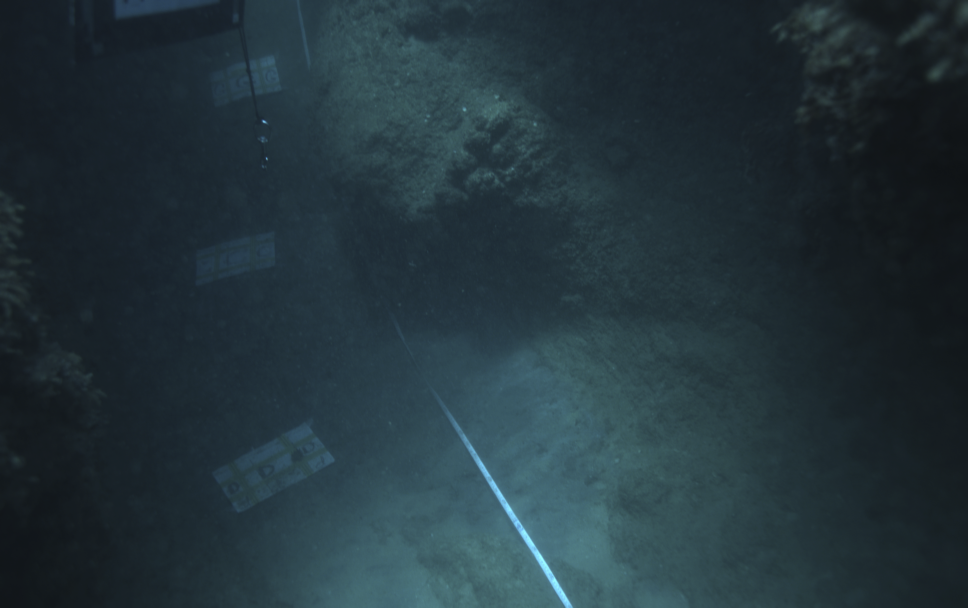}%
    \end{minipage}
    \caption{Optical challenges in underwater imagery. Attenuation (left), reflections (middle) and turbidity (right).}
    \label{fig:underwater_challenges}
\end{figure}

Autonomous underwater vehicles often rely on visual based sensing for localization. For these systems, it is generally sufficient to build a sparse representation of the scene, using landmarks or detected feature points to optimize the vehicle pose estimates~\cite{slam_survey}. However, some tasks require a vehicle to operate near or interact with the environment, such as conducting close range imaging surveys or performing manipulation tasks. For such applications, a dense and real-time understanding of the scene is required for safe autonomous operation of the vehicle to enable obstacle avoidance and intervention planning. A practical solution to this problem must be robust to underwater imaging effects (Fig.~\ref{fig:underwater_challenges}), such as nonlinear attenuation, backscatter, and inconsistent lighting, must provide metrically accurate depth measurements, must be computationally efficient for real-time deployment on embedded compute systems, and ideally should be compact and low-cost for integration on diverse underwater robotic platforms.

Active-light depth sensors, such as LiDAR \cite{uw_lidar_inspection} or RGB-D cameras \cite{uw_kinect,uw_depthcam}, are commonly used in terrestrial applications and provide an intuitive solution for metrically accurate depth sensing. However, these sensors are challenging to deploy underwater, due to nonlinear, frequency dependent attenuation of light in the water column, refraction effects, and the scattering effects of suspended particulates. Imaging sonars provide an alternative solution that is robust to underwater lighting effects, and they are used in many marine field robotics applications~\cite{s21237849}. However, they are expensive and provide limited spatial resolution compared to optical sensors with noisy depth estimates that make them less suitable for complex intervention and close range inspection tasks.



Passive stereo cameras provide a low-cost, off-the-shelf solution that is simple to setup and can provide dense, high-resolution depth measurements of the scene. However, stereo cameras can be bulky to integrate in underwater housings with an adequate camera baseline, making them difficult to deploy on small underwater systems or on the front of streamlined AUVs with highly limited space. Passive monocular cameras provide an attractive alternative to stereo cameras, due to their compactness and being readily integrated on any underwater platform. For this reason, predicting dense depth from monocular images for marine robotics has gained more attention in recent years, with significant advancements made in deep learning approaches.

In this work, we extend the recent advancements in learning based monocular depth estimation to demonstrate a practical approach for metrically accurate depth predictions that is suitable for deployment on diverse underwater platforms and is transferable to different camera systems without retraining, special calibration, or parameter tuning. In particular, we make the following contributions:
\begin{itemize}
    \item An extension of a state-of-the-art lightweight network for dense depth prediction from monocular images to fuse triangulated feature points into the prediction stages through a dense parameterization of these sparse priors, improving the prediction accuracy and providing metric scale constraints.
    \item Training and evaluation of the method on the FLSea~\cite{flsea} dataset, demonstrating significantly improved performance through fusion of the sparse priors.
    \item Evaluation of the method on an additional image dataset from a coral reef survey we collected off of Lizard Island, Australia, with a downward facing diver operated camera rig. No images from this dataset or camera rig were used in the training of the method, demonstrating that the method generalizes well to different camera systems and environments.
\end{itemize}


\section{Related Work}
\label{sec:related_work}

\subsection{Monocular Underwater Depth Estimation}

Some early approaches to underwater depth estimation from monocular images were driven by the need to enhance underwater images~\cite{uw_vision_enhancement}. These methods aimed to mitigate the challenges posed by underwater lighting and attenuation effects, effectively transforming underwater images to resemble those taken in a terrestrial environment. To achieve this, researchers constructed models that explicitly account for the physics underlying underwater illumination and image formation. These methods, known as image formation models, incorporate coefficients to represent attenuation and other relevant water column properties. More recently, supervised deep learning approaches have shown great potential for learning image formation models and predicting depth maps from monocular images~\cite{uw_gan}. However, image formation models suffer from poorly constrained depth estimates and sensitivity to difficult imaging conditions such as turbidity and reflections.

Given the challenges and expense of collecting large-scale underwater datasets in diverse environments for training supervised models, another line of research has focused on generating realistic synthetic underwater datasets to train depth prediction models without the need for real images~\cite{uw_synthesis1, uw_synthesis2}. While a promising avenue of research, the synthesis of underwater data is not yet able to fully capture the realistic formation of underwater images, leading to degraded performance on real data.

Recent work has achieved state-of-the-art performance through transfer learning, where a monocular depth estimation model is trained on large-scale terrestrial data and then fine tuned with a smaller underwater dataset. \cite{udepth} provides a survey of these recent works, where popular adopted models include Monodepth2~\cite{monodepth2}, AdaBins~\cite{adabins}, and U-Net~\cite{u_net_review} based architectures. Additionally, by incorporating domain specific loss formulations, the performance of the underlying models can be further improved for underwater environments~\cite{udepth}. Our method extends the work of AdaBins~\cite{adabins} and UDepth~\cite{udepth} in order to boost the prediction accuracy and provide metrically accurate scale through the fusion of sparse feature priors in the network architecture.



\subsection{Sparse Measurement Fusion}
An inherent problem of monocular camera systems is scale ambiguity. While supervised deep learning models are able to predict metric depth from monocular RGB image inputs without additional measurements, such models suffer from poor generalizability when the data distribution of the real world application does not match the distribution of the training data or the camera parameters change. Sparse depth priors from external measurements can be used to deliver a valuable depth prior and guidance signal for estimation in accurate metric scale, as surveyed in~\cite{depth_compl_survey}. We provide a brief review here of learning based approaches for sparse prior fusion in depth prediction.

Categorized as \textit{early fusion}, the first learning based methods simply concatenated the sparse depth maps and RGB image together as the network input~\cite{early_fusion_1}, or convolved them in the first few layers of the model. In recent works, \textit{late fusion} techniques have been developed with more complex architectures such as dual-encoder~\cite{dual_encoder} or double encoder-decoder~\cite{double_enc_dec} models to better encode the sparse depth prior into the RGB image feature space. As a result, late fusion techniques generally outperform early fusion models at the cost of increased complexity. To cope with the high sparsity of the depth input, these methods often use either binary or continuous validity masks to guide the convolution operations to the relevant parts of the input prior. In the most recent approaches, researchers have begun to incorporate surface normal representations~\cite{lidar_fusion_early_normal_binary}, affinity~\cite{affinity}, and residual depth maps~\cite{residual_depth} into early, late or hybrid fusion architectures.

Many deep learning based methods for depth estimation from monocular images treat their problem in a one-shot fashion. At each inference step a new image is fed into the network to predict depth with no connection between subsequent frames or timesteps. In robotics however, these image inputs usually come from video streams, where there is a large correlation between sequential frames. Notably, many intervention robots are already capable of accurate localization through off-the-shelf  visual-inertial methods like ORB-SLAM3 \cite{orbslam3}, which store knowledge about the environment in the form of sparse triangulated visual feature points. As proposed in methods like \cite{depth_from_rgb_and_sparse}, such sparse depth priors can be used in fusion with corresponding RGB images to provide metric scale cues to a learning method, solving the problem of scale ambiguity and improving the inference accuracy. We draw inspiration from these prior works in our approach to fusing sparse feature based priors into the depth prediction network architecture, following a late fusion approach.
\section{Method}
\label{sec:method}

\subsection{Depth Prior Parameterization}
\label{ssec:depth_prior_parameterization}

To overcome the problem of scale ambiguity and to recover the global scale of the scene, we develop an approach to include sparse depth measurements of the scene in the form of visual features as input to the depth prediction network. We adapt the densified parameterization suggested by \cite{depth_from_rgb_and_sparse}, which enables fusion of sparse priors with arbitrary density.

\subsubsection{Sparse Prior Generation}

By using the video stream from a robotic system, we can track and triangulate visual keypoints between frames. Popular visual-inertial SLAM methods like ORB-SLAM3 \cite{orbslam3} track and triangulate features over many frames, resulting in highly accurate priors.
As a simplification for our evaluation on the FLSea \cite{flsea} dataset, we use the matched visual features between two consecutive frames and extract the depth value from the available dataset ground truth. These sparse measurements are then used as the prior guidance signal for the depth prediction task. The steps to generate a sparse set of prior depth measurements are as follows:
\begin{enumerate}
    \item The image is divided into equal sized patches and a fixed number of SIFT keypoints~\cite{Lowe2004DistinctiveIF} are detected in each patch.
    \item The detected features are matched between frames by performing bidirectional nearest neighbor search followed by enforcing an epipolar constraint for outlier filtering. The epipolar constraint can be described as $k_i^1 = F^T k_i^2$, where $(k_i^1, k_i^2)$ denotes the $i$-th keypoint match between frames $(1,2)$ and $F$ is the fundamental matrix.
    \item For each inlier feature point, depth values are extracted from the ground truth depth maps at the keypoint location.
\end{enumerate}


For evaluation on the LizardIsland dataset, we use the SLAM method developed in~\cite{billings2022hybrid} to generate feature priors from the image sequence, where the feature positions and depths are projected from the constructed SLAM feature map, demonstrating the practical performance of the method as it would be deployed on an underwater vehicle with an online SLAM system.

\subsubsection{Dense Prior Parameterization}

Since the amount of available keypoints may vary from image to image, a generalized representation of constant size and suitable format is desirable to feed into the network.
Inspired by \cite{depth_from_rgb_and_sparse}, we define two maps $S_1(x,y)$ and $S_2(x,y)$ to map the sparse set of keypoints $k_i$ into two continuous image representations.

Following the approach of \cite{depth_from_rgb_and_sparse}, $S_1$ is a nearest neighbor interpolation of the sparse keypoint depths into a full size dense depth map, where every pixel $(x,y)$ in $S_1(x,y)$ takes the depth value $d_{k^*}$ of its closest keypoint $k^*$. The formula to compute $S_1$ is given as:

\begin{equation}
    S_1(x,y) = d_{k^*} \text{, where } k^* = \underset{i}{\arg\min}||(x,y)-(x_i,y_i)||_2
\label{eq:S1}
\end{equation}

$S_2$ is formulated as a probability map to account for the distance of each pixel to its nearest keypoint. Let $r(x,y|k_i)$ be the euclidean pixel distance of point $(x,y)$ to keypoint $k_i$. We assume the probability with regards to the pixel distance to a given keypoint to follow a normal distribution. The computation of $S_2$ then follows as:

\begin{equation}
    S_2(x,y) = \frac{1}{\sigma \sqrt{2\pi}} \exp \left(-\frac{r(x,y|k^*)^2}{2\sigma^2} \right),
\label{eq:S2}
\end{equation}
where $r(x,y|k^*) = \underset{i}{\min}||(x,y)-(x_i,y_i)||_2$ and $\sigma$ is the standard deviation which is treated as a tuning parameter. As a result of empirical testing, we use $\sigma=10$ for all our experiments.

\begin{figure}[!t]
    \centering
    \includegraphics[width=0.25\linewidth]{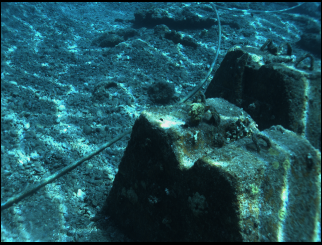}%
    \includegraphics[width=0.25\linewidth]{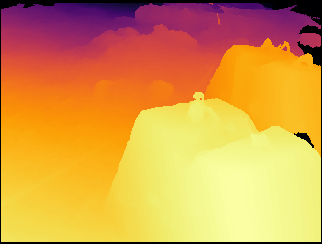}%
    \includegraphics[width=0.25\linewidth]{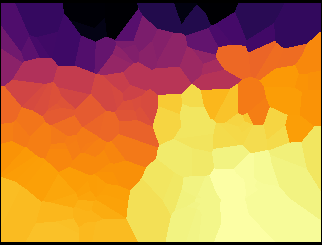}%
    \includegraphics[width=0.25\linewidth]{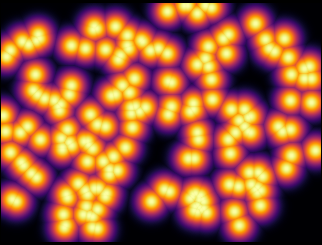}%
    \caption{Dense prior parameterization. RGB (first), ground truth depth data available from the given dataset (second), the $S_1$ nearest neighbor depth map (third) and the $S_2$ prior probability map (fourth), where the probability is highest (bright spots) at the location of the features.}
    \label{fig:dense_prior_parameterization}
\end{figure}

To form the prior input to our model, the two maps $S_1$ and $S_2$ are depth-wise concatenated in the form of a two-channel image. This image is then fed into the network at multiple stages. Figure \ref{fig:dense_prior_parameterization} shows an example of the parameterized prior inputs.

\subsection{Architecture Design}
\label{ssec:architecture_design}
\begin{figure}[!t]
    \centering
    \includegraphics[width=1.0\linewidth]{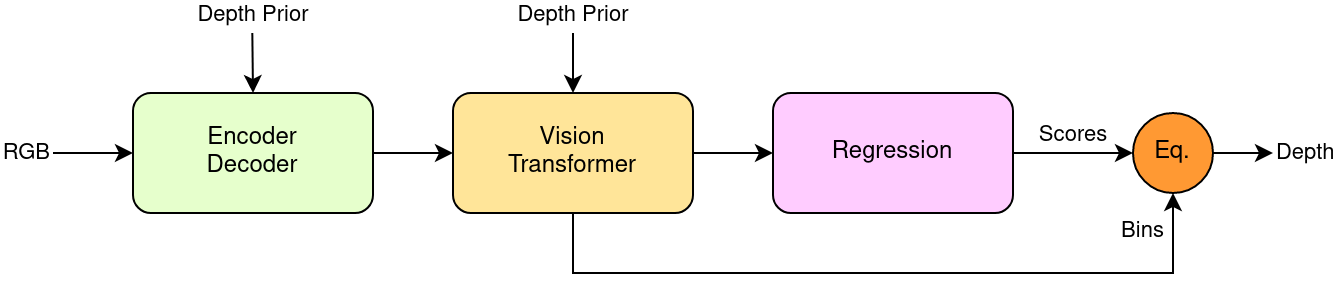}
    \caption{Model overview with three main components based on AdaBins \cite{adabins}: Encoder-decoder, vision transformer and regression. The depth prior is fed into the model at multiple stages during the decoder as well as in the beginning of the vision transformer.}
    \label{fig:model_overview}
\end{figure}

The proposed network architecture is based on the popular AdaBins \cite{adabins} model and the contributions of UDepth \cite{udepth}. 
The architecture consists of a lightweight encoder-decoder backbone, an efficient vision transformer and a convolutional regression module. Figure \ref{fig:model_overview} shows an overview of the model architecture.


\subsubsection{Encoder Decoder Backbone}
In the first module of the proposed network, we use a computationally fast encoder-decoder backbone.
Motivated by the contributions of UDepth~\cite{udepth}, we exchange the heavier encoder module of AdaBins~\cite{adabins} with a more efficient model based on MobileNetV2~\cite{mobilenetv2}. This model is optimized for mobile architectures and lays the foundation for many applications which need real-time performance. Additionally, UDepth has demonstrated that this architecture attains good feature representations for underwater applications.
The encoder-decoder module is constructed in a U-Net fashion~\cite{u_net_review}: In the encoder, we pass the input RGB image through all layers of the MobileNetV2 model and extract the signal at multiple stages for later use as skip connections. The decoder then consists of several upsampling layers which combine the current signal with skip connections from the encoder layers combined with the appropriately scaled depth parameterization. 

\subsubsection{Vision Transformer}

While the preceding encoder-decoder block provides spatial feature information at multiple resolutions, the vision transformer component refines the results with global context. 
\begin{figure}[!t]
    \centering
    \includegraphics[width=1.0\linewidth]{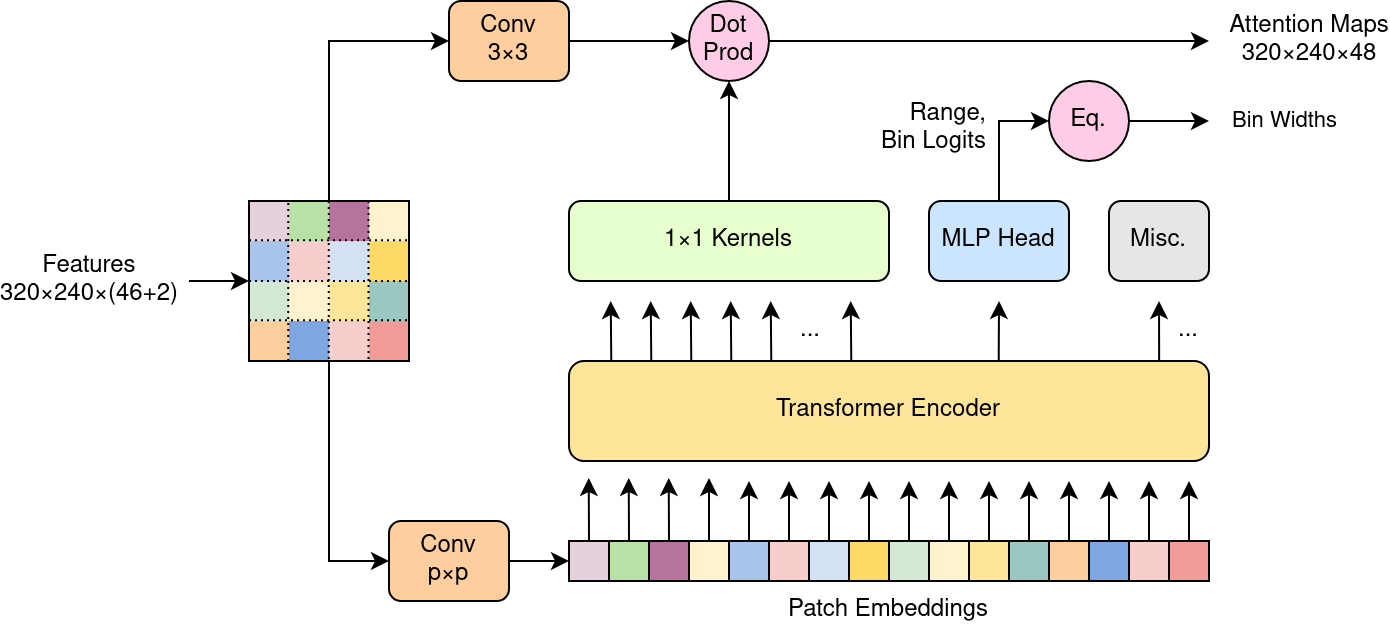}
    \caption{Vision Transformer Structure based on AdaBins \cite{adabins}. The feature image coming from the preceeding encoder-decoder module and depth parameterization are spatially divided into a number of patches of size $p \times p$. Using a  convolutional block we have a set of patch embeddings as input sequence for the transformer encoder. After the encoder, one output embedding is used as input for a multilayer perceptron (MLP) head to predict the bin width logits and additionally range $r$, from which we then compute the metric bin widths via Equation \ref{eq:b_i}. A longer subset of output embeddings is used as attention query kernels for forming the output attention maps via dot product.}
    \label{fig:vision_transformer}
\end{figure}
While vision transformers in general are computationally expensive~\cite{vit_survey}, AdaBins introduced a lightweight vision transformer that we use with some modifications~\cite{adabins}.
Instead of estimating absolute depth values directly, they divide the depth range into $n$ bins with adaptive size and predict the Softmax classification score for every pixel to belong to each of the bins. 
While we adopt this mechanism for estimating adaptive bin widths and range attention maps, our modified architecture additionally predicts the depth range $r$ for each image. That is, we estimate the bin widths $b_i$ by multiplying the normalized bin widths by the estimated range rather than a fixed upper bound, as was implemented in~\cite{adabins}. In our testing, we found that this modification to the network improves performance on inputs which are at different ranges than the training data.

To obtain the normalized bin widths $\Bar{b}_i$, the adaptive bin logits $\Tilde{b}_i$ output by the transformer are normalized such that they sum up to 1. For strictly positive bin widths, a small positive value $\epsilon =10^{-3}$ is added to all logits. The formulation for computing the bin widths $b_i$ is as follows:

\begin{equation}
    b_i = r \cdot \Bar{b}_i = r \cdot \frac{\Tilde{b}_i + \epsilon}{\sum_j^n(\Tilde{b}_j + \epsilon)} \text{ , where } i = 1,2, ..., n
    \label{eq:b_i}
\end{equation}

The bin width logits $\Tilde{b}_i$ and range $r$ are predicted through a multilayer perceptron (MLP) head with $n+1$ outputs ($n$ bins plus 1 for range $r$) from the first transformer output embedding.
Figure \ref{fig:vision_transformer} shows a visualization of our modified vision transformer module.



\subsubsection{Convolutional Regression}
The last stage of the network forms the depth image prediction from the bin centers and range attention maps output by the vision transformer module. 
The bin classification scores are computed following the approach of AdaBins~\cite{adabins}, where the range attention maps from the vision transformer module are $1\times1$ convolved, followed by a Softmax activation to map the output logits into a set of $n$ probabilities for each pixel, where $n$ is the number of bins. The final depth image prediction $\Hat{d}(x,y)$ is then formed as the linear combination of the probabilities $p_i(x,y)$ and bin centers $c_i$, 
as formulated in Equation \ref{eq:linear_combination}:

\begin{equation}
    \Hat{d}(x,y) = \sum_i^n c_i \cdot p_i(x,y)
    \label{eq:linear_combination}
\end{equation}


\subsection{Loss Functions}
\label{ssec:loss_functions}

The training loss is formulated as a combination of different loss functions, providing different constraints on the network predictions. In the following, we denote $\hat{d}$ as the predicted depth and $d$ as the available ground truth.

\subsubsection{Root Mean Squared Error}
The traditional RMSE loss function directs the network to learn an exact metric scale prediction. This loss is formulated as

\begin{equation}
    \mathcal{L}_{\text{RMSE}}(\Hat{d}, d) = \sqrt{\frac{1}{N}\sum_i^N(\Hat{d}_i-d_i)^2}
    \label{eq:rmse}
\end{equation}

\subsubsection{Scale Invariant Logarithmic Loss}
For robotic intervention tasks errors are often calculated in logarithmic space \cite{mono_d_estim_overview}, which penalizes errors at close range and becomes more forgiving at greater distances.
Motivated by \cite{adabins, udepth}, we utilize a parameterized variation of the Scale Invariant Logarithmic Loss (SILog Loss)\cite{silog}, which allows us to balance the learning focus between estimating 
accurate metric scale
and relative depth:
\begin{equation}
    \mathcal{L}_{\text{SILog}}(\Hat{d}, d) = \beta \sqrt{ \frac{1}{N}\sum_i^N g_i^2 - \frac{\lambda}{N^2}(\sum_i^N g_i)^2 } 
    \label{eq:silog_param}
\end{equation}
where $g_i = \log \Hat{d}_i - \log d_i$ and $\lambda \in [0,1]$. As proposed by~\cite{adabins}, we use $\lambda=0.85$ and $\beta = 10$ in all of our experiments.

\subsubsection{Chamfer Distance Loss}

To guide the estimation of adaptive bin sizes $b_i$ (described in Subsection \ref{ssec:architecture_design}), we utilize a regularizer which encourages the distribution of bin centers to follow the actual depth distributions of a given input image. Following~\cite{adabins}, we use the Chamfer Distance Loss first introduced by~\cite{chamfer}:

\begin{equation}
    \begin{aligned}
        \mathcal{L}_{\text{Chamfer}}(c, d) = \sum_{c_i \in c} \min_{d_j \in d}||d_j - c_i||_2^2 + \sum_{d_j \in d} \min_{c_i \in c}||d_j - c_i||_2^2 \:, &\\
        \text{where } i=1,2,...,n, \; j=1,2,...,N &
    \end{aligned}
    \label{eq:chamfer}
\end{equation}

\subsubsection{Learning Objective}

As our learning objective, we use a weighted combination of the losses $\mathcal{L}_{\text{RMSE}}$, $\mathcal{L}_{\text{SILog}}$ and $\mathcal{L}_{\text{Chamfer}}$, with $\lambda_1 = 0.3$, $\lambda_2 = 0.6$ and $\lambda_3 = 0.1$.

\begin{equation}
    \mathcal{L}_{\text{Objective}} = \lambda_1 \cdot \mathcal{L}_{\text{RMSE}} + \lambda_2 \cdot \mathcal{L}_{\text{SILog}} + \lambda_3 \cdot \mathcal{L}_{\text{Chamfer}}
\end{equation}

\section{Experiments}
\label{sec:experiments}


\subsection{Datasets}
\label{ssec:datasets}

\subsubsection{FLSea}

Our method is trained and evaluated on the FLSea dataset~\cite{flsea}, which features forward looking imagery close to the sea floor in relatively shallow water depths less than $\SI{10}{\meter}$. On average, the maximum distance range in the available ground truth is around $\SI{10}{m}$. 
The publicly available dataset consists of 22'451 image frames with RGB and metric ground truth depth data at 12 different underwater locations. Of these 12 locations, we use 10 for training and 2 for testing (\textit{u\_canyon} and \textit{sub\_pier}). FLSea showcases many of the lighting challenges described in Section \ref{sec:related_work}. In many instances, the imagery includes reflections, turbidity and attenuation all at the same time. Despite providing color corrected images, we use the raw images both during training as well as testing. Example pairs of RGB and depth map data from the FLSea dataset are shown in Figure~\ref{fig:results_FLSea}.



Obtaining true underwater depth is a very challenging task. As explained in Section~\ref{sec:introduction}, sensors such as LiDAR or RGB-D cameras have only very limited performance due to the physical properties of the water column. For this reason, the ground truth depth maps of FLSea are generated 
using the Agisoft Metashape photogrammetry software~\cite{metashape}. Photogrammetry pipelines such as Metashape are considered state-of-art solutions for underwater 3D reconstruction in both research and industry. However, due to their extreme computational cost, such pipelines can not be deployed in online applications. Achieving offline photogrammetry accuracy with a real-time method is an ultimate aim of our research.


In a pre-processing step, we match features between frames and extract sparse prior points with their depth values for all images in the FLSea dataset. For this, we follow the procedure described in Subection \ref{ssec:depth_prior_parameterization} and store the keypoints as a list for later usage during training and testing.

\subsubsection{LizardIsland Dataset}

In order to test the proposed method's generalization to different camera systems and underwater environments not represented in the training data, we additionally evaluate the method performance on a dataset collected with a downward facing diver operated camera rig conducting a survey of a Coral Reef off of Lizard Island, Australia. The dataset is composed of a stereo image sequence which was processed through COLMAP~\cite{schoenberger2016sfm} to obtain ground truth dense depth maps. The sequence was also processed through the SLAM method published in~\cite{billings2022hybrid} to generate optimized feature maps that were projected back into the left camera image frame to obtain the sparse depth priors for input into the network. Example images from this dataset along with the sparse prior visualizations are shown with the evaluation results in Figure~\ref{fig:lizard}, where the left camera image was used as the monocular input to the network. The survey dataset includes 2'216 sequential images that were included in the evaluation of the method performance.

\subsection{Learning Pipeline}
For the implementation of this work we used the PyTorch~\cite{pytorch} library. We initialize our model at random, except for the MobileNetV2~\cite{mobilenetv2} part in our encoder-decoder module (see Subsection \ref{ssec:architecture_design}). As our optimizer we choose AdamW~\cite{adamw} and train all 15.6~M network parameters freely in steps with a batch size of 6. 
For the learning rate $lr(t) = lr_0 \cdot r^t$ we use exponential decay over time, 
where $lr_0=0.0001$ is the base learning rate and $r=0.9$ is the decay rate.
The model is trained in a supervised fashion on training data from FLSea.
For training robustness, we apply a series of random transforms to our inputs: Horizontal flips, color- and brightness scaling, as well as depth scaling. 
Throughout all training for these tests, we randomly select 200 depth priors per image frame. 
From these sparse priors we form our dense prior parameterization (see Subsection~\ref{ssec:depth_prior_parameterization}) and feed it into the network as illustrated in Figure~\ref{fig:model_overview}. Note that with an output resolution of $320\times240$, the 200 keypoints are extremely sparse and only cover 0.26\% of the total number of pixels. The training loss is then computed between the prediction outputs and target depth maps by using the learning objective formulated in Subsection~\ref{ssec:loss_functions}.

Using a NVIDIA GTX 2080 graphics card, the pipeline training took approximately 1 hour per epoch. The best results were achieved after around 10 hours of training.

\subsection{Evaluation Metrics}

To evaluate the performance of the trained models, we use a variety of standard error metrics. The error metrics between the predicted $\Hat{d}$ and ground truth depth $d$ are calculated as follows:
For its straightforward interpretation, we choose standard RMSE in linear space (Eq. \ref{eq:eval_rmse_lin}).
Following the intuition in Subsection \ref{ssec:loss_functions}, we further evaluate on RMSE in logarithmic space (Eq. \ref{eq:eval_rmse_log}) to investigate emphasized errors at close range. We also compare this error against its scale invariant counterpart (Eq. \ref{eq:eval_rmse_silog})
to analyze the accuracy of the overall estimated scene scale.
\begin{equation}
    \text{RMSE}_{\text{lin}}(\Hat{d}, d) = \sqrt{ \frac{1}{N}\sum_i^N (\Hat{d}_i - d_i)^2}\\
    \label{eq:eval_rmse_lin}
\end{equation}
\begin{equation}
    \text{RMSE}_{\text{log}}(\Hat{d}, d) = \sqrt{ \frac{1}{N}\sum_i^N (\log \Hat{d}_i - \log d_i)^2}
        \label{eq:eval_rmse_log}
\end{equation}
\begin{equation}
    \text{RMSE}_{\text{silog}}(\Hat{d}, d) = \sqrt{ \frac{1}{N}\sum_i^N (\log \Hat{d}_i - \log d_i + \alpha(\Hat{d}, d))^2},
    \label{eq:eval_rmse_silog}
\end{equation}
where $\alpha(\Hat{d}, d) = \frac{1}{N}\sum_i^N (\log d_i - \log \Hat{d}_i)$ is the term that makes the error scale invariant \cite{silog}.

Furthermore, we analyze the Mean Absolute Relative Error to gain insight into the relative accuracy of our prediction:

\begin{equation}
    \text{MARE}(\Hat{d}, d) = \frac{1}{N} \sum_i^N \frac{|\Hat{d}_i - d_i|}{|d_i|}
\end{equation}

\subsection{Results}
\label{ssec:results}

\subsubsection{FLSea}

\begin{figure}[!t]
    \centering
    \includegraphics[width=\linewidth]{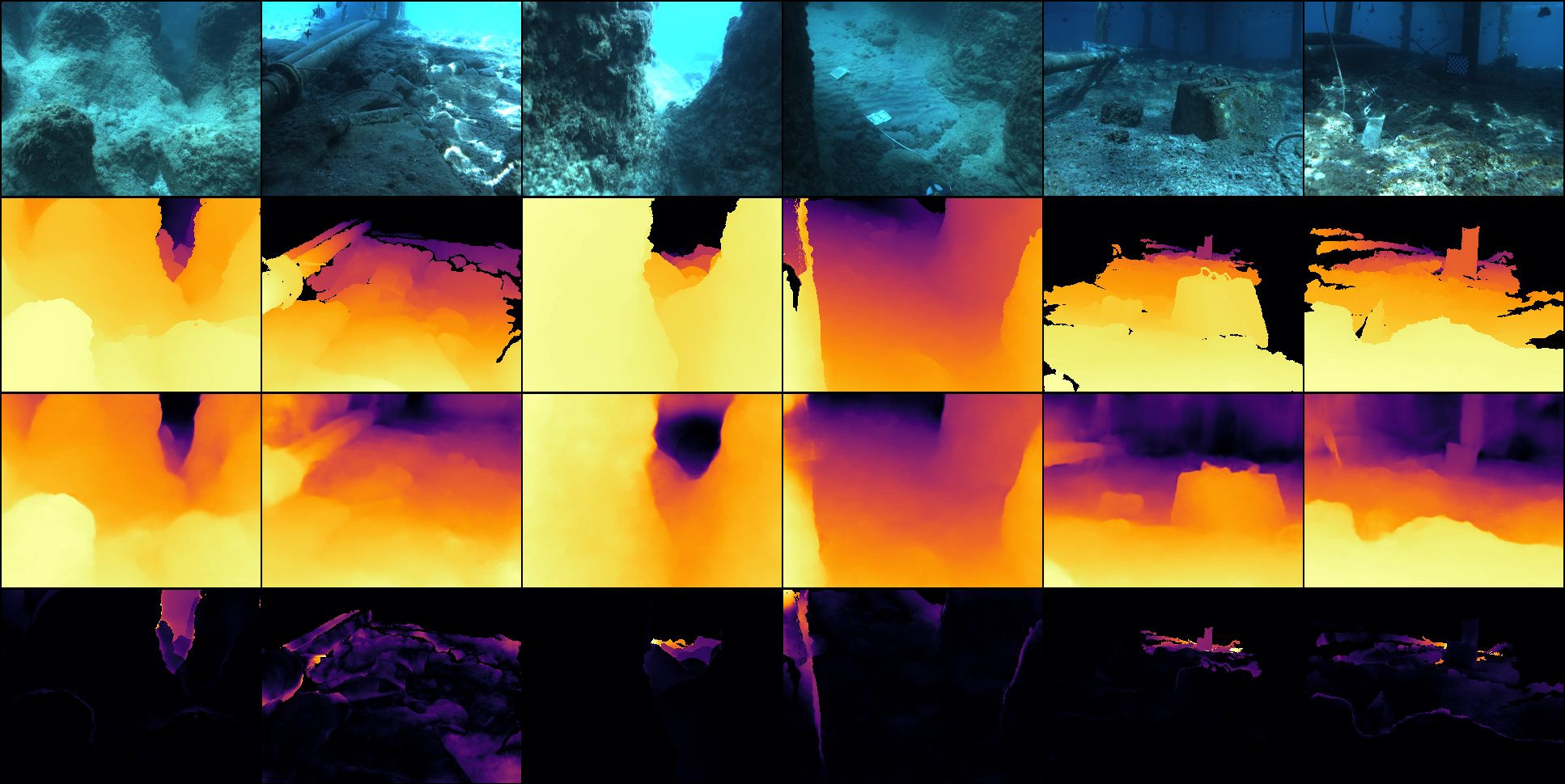}
    \caption{Depth prediction examples on the FLSea dataset. RGB input (first row), ground truth depth map (second row), predicted depth map (third row) and absolute error (fourth row). Despite various optical challenges such as attenuation, reflections and turbidity our approach yields stable predictions.}
    \label{fig:results_FLSea}
\end{figure}

We evaluate the performance of the network on the test split of the FLSea~\cite{flsea} dataset and analyze how the prediction improves by using sparse priors. As when training, we use 200 randomly selected sparse priors per image which are stored from our pre-processing step. Figure~\ref{fig:results_FLSea} shows a visualization of our predictions and Table~\ref{tab:flsea_results} presents the evaluated error metrics, cut off at different ranges to provide insight into near depth and far depth prediction accuracy.

\begin{table}
\centering
\caption{Evaluation metrics of the network tested on our test split of the FLSea dataset. The usage of sparse priors versus without shows significant improvements all across the board. When using sparse priors, the metrics for RMSE in log versus silog space are very similar, indicating a good estimation of our predicted depth range.}
\label{tab:flsea_results}
\begin{tabular}{ll|rrrr}

\textbf{Range}           & \textbf{Priors}         & \textbf{RMSE}          & \textbf{RMSE}          & \textbf{RMSE}          & \textbf{MARE}         \\
                &                & \textbf{(linear)}      & \textbf{(Log) }        & \textbf{(SILog)}       &              \\
\hline
\hline
Full range      & 0              & 0.596         & 0.171         & 0.127         & 0.137        \\
                & 200            & 0.372         & 0.089         & 0.087         & 0.037        \\
\cline{3-6}
                & Improvement    &  37.6\%       &  48.0\%        &  31.5\%        &  73.0\%       \\
\hline
$d < \SI{5}{m}$ & 0             & 0.390         & 0.157         & 0.102         & 0.134         \\
                & 200           & 0.167         & 0.057         & 0.055         & 0.031         \\
\cline{3-6}
                & Improvement   &  57.2\%       &  63.7\%        &  46.1\%        &  76.9\%       \\

\hline
$d < \SI{1}{m}$ & 0             & 0.263         & 0.254         & 0.049         & 0.281         \\
                & 200           & 0.040         & 0.042         & 0.032         & 0.029         \\
\cline{3-6}

                & Improvement   &  84.8\%       &  83.5\%        &  34.7\%        &  89.7\%       \\

\end{tabular}
\end{table}

The results of Table~\ref{tab:flsea_results} demonstrate significantly improved depth prediction accuracy for all depth ranges and across all evaluation metrics through the fusion of sparse feature priors into the network. 
It is important to reiterate that the ground truth depth maps of the FLSea~\cite{flsea} dataset were computed through a photogrammetric 3D reconstruction pipeline, so these metrics represent the ability of our method to reproduce these photogrammetry results. However, as discussed in Subsection~\ref{ssec:datasets}, such pipelines are recognised as state-of-art solutions for both research and industry.

\subsubsection{LizardIsland}

\begin{figure}[t]
    \centering
    \includegraphics[width=\linewidth]{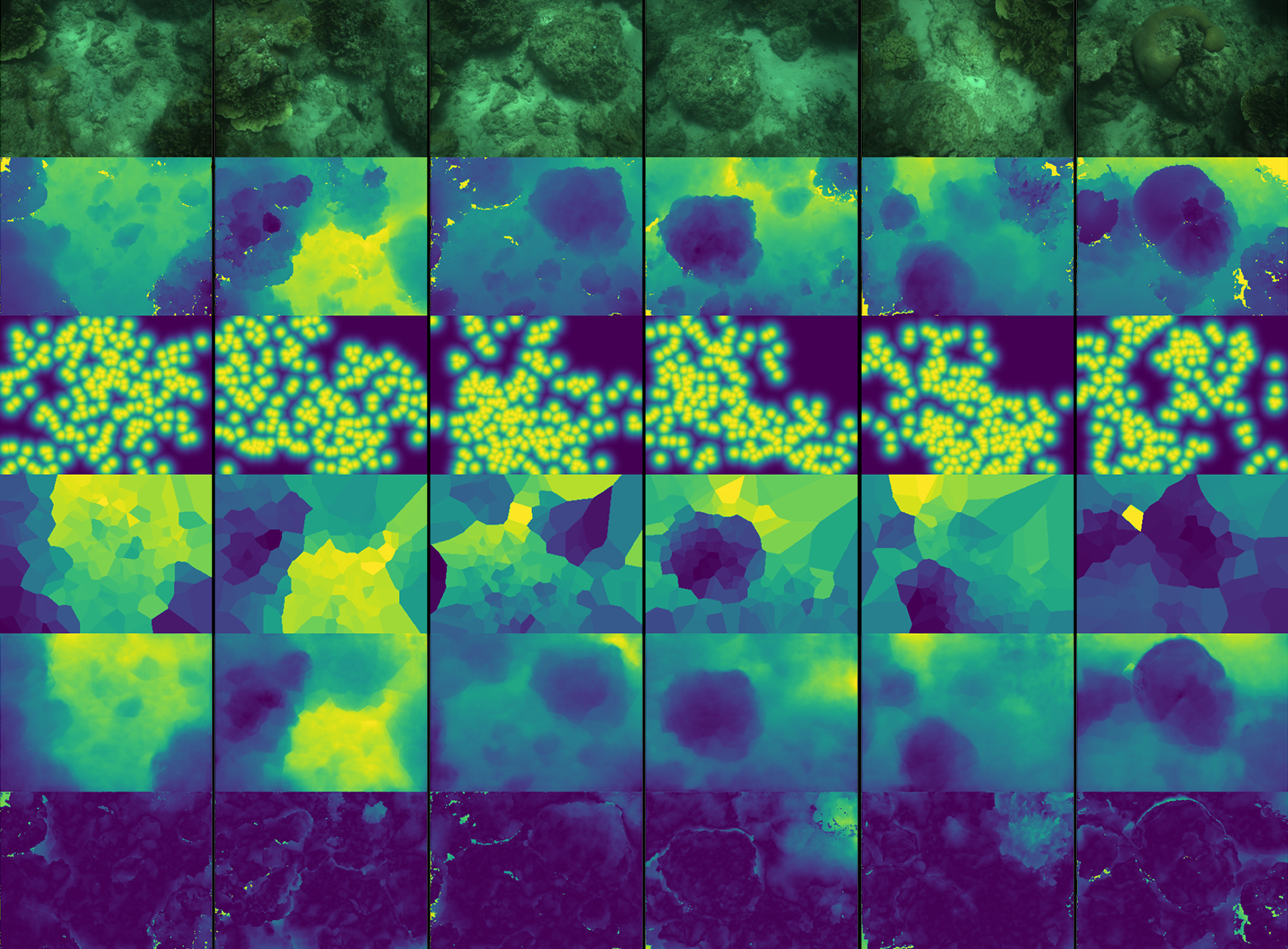}
    \caption{Visualization of prediction results on samples from the LizardIsland dataset. 
    RGB input (first row), ground truth depth map (second row), prior probability distribution map (third row), prior nearest neighbor map (fourth row), predicted depth map (fifth row) and absolute error between the prediction and ground truth depth maps (sixth row).}
    \label{fig:lizard}
\end{figure}

We evaluated the performance of the method on the LizardIsland dataset, with the model trained only on the FLSea dataset. Figure~\ref{fig:lizard} shows visualizations of the predictions for a few samples from this dataset, and Table~\ref{tab:lizard} gives the quantitative evaluation of the depth prediction accuracy. In addition to evaluating the predicted depth map accuracy, we also evaluate the accuracy of the nearest neighbor prior depth maps, using the same evaluation metrics. The poor accuracy of the raw nearest neighbor prior maps indicates that they are too coarse of an estimate to be useful for dense depth prediction by themselves. However, the network predictions achieve similar accuracy on the LizardIsland sequence as they do on the test sequences of the FLSea dataset, demonstrating that the method can effectively generalize to novel underwater environments and camera systems not included in the training data. 

\begin{table}
\centering
\caption{Evaluation of the proposed method on the LizardIsland dataset, with the model trained only on the FLSea~\cite{flsea} dataset. The evaluated depth range was limited to \SI{5}{\meter}. 
Mean accuracy of the predicted depth maps (top row) versus standalone input nearest neighbor prior (bottom row).}
\label{tab:lizard}
\begin{tabular}{lcccc}
& \multicolumn{1}{c}{\textbf{\begin{tabular}[c]{@{}c@{}}RMSE\\ (linear)\end{tabular}}} & \textbf{\begin{tabular}[c]{@{}c@{}}RMSE\\ (log)\end{tabular}} & \multicolumn{1}{c}{\textbf{\begin{tabular}[c]{@{}c@{}}RMSE\\ (SIlog)\end{tabular}}} & \multicolumn{1}{c}{\textbf{\begin{tabular}[c]{@{}c@{}}MARE\\ \phantom{}\end{tabular}}} \\ \hline
\hline
Prediction & 0.0876 & 0.0430 & 0.0410 & 0.0274 \\
Prior & 1.4790 & 3.9999 & 3.1571 & 0.7230 \\ 
\end{tabular}
\end{table}
\section{Conclusion}

In this work, we have demonstrated a compact and cost-efficient dense depth sensing solution for mobile underwater robots. Our method follows a supervised monocular deep learning approach, where we fuse sparse depth measurements in order to solve the problem of scale ambiguity and improve the depth prediction accuracy. Experiments conducted on challenging underwater datasets demonstrate significantly improved depth prediction performance through fusion of the sparse priors.
 With a Root Mean Squared Error of $\SI{0.167}{\meter}$ ($\SI{5}{\meter}$ range) and $\SI{0.040}{\meter}$ ($\SI{1}{\meter}$ range) on the FLSea~\cite{flsea} dataset, we achieve promising accuracies for using this method in underwater intervention and close range survey applications. However, some intervention tasks may require higher levels of precision, motivating further research to improve this method. Considering that our approach requires only a monocular camera and some knowledge about the robot's motion, our solution is easy to integrate and particularly attractive for low-cost and small sized vehicles. Regarding computational efficiency, our proposed model remains lightweight and achieves real-time framerates at $\SI{160}{\Hz}$ on a NVIDIA GTX 2080 gpu or $\SI{7}{\Hz}$ on an Intel i9-9900K cpu using a single core.

In an outlook for future work, we will explore replacing the naive nearest neighbor approach of the depth prior parameterization with a smoother representation or better informed guess, such as propagating the previous prediction as an input for the current prior guess. In a similar matter, we would seek to integrate knowledge about the sparse depth priors' uncertainty which is often available when using SLAM pipelines. Furthermore, we would like to incorporate a segmentation for points at infinity as a side task to handle such areas more robustly.

\bibliographystyle{IEEEtran}
\bibliography{bibliography/references}

\end{document}